\documentclass[runningheads]{llncs}
\usepackage{graphicx}

\usepackage{tikz}
\usepackage{comment}
\usepackage{amsmath,amssymb} 
\usepackage{color}

\usepackage[activate={true,nocompatibility},final,tracking=true,kerning=true,spacing=true,factor=1100,stretch=10,shrink=9]{microtype}
\usepackage{times}
\usepackage{epsfig}
\usepackage{graphicx}
\usepackage{amsmath}
\usepackage{amssymb}
\usepackage{adjustbox}
\usepackage{float}

\usepackage[pagebackref=true,breaklinks=true,colorlinks,bookmarks=false]{hyperref}

\usepackage{xcolor}
\definecolor{RED}{rgb}{1, 0, 0}
\definecolor{fig_red}{rgb}{1, 0, 0}
\definecolor{fig_green}{rgb}{0, 0.6, 0}
\newcommand{\squeezeup}{\vspace{-2mm}}

\begin{document}
\pagestyle{headings}
\mainmatter
\def\ECCVSubNumber{4976}  

\title{StyleBabel: Artistic Style Tagging and Captioning} 


\titlerunning{StyleBabel: Artistic Style Tagging and Captioning}
%
\author{Dan Ruta\inst{1} \and
Andrew Gilbert\inst{1} \and
Pranav Aggarwal\inst{2} \and
Naveen Marri\inst{2} \and
Ajinkya Kale\inst{2} \and
Jo Briggs\inst{3} \and
Chris Speed\inst{4} \and
Hailin Jin\inst{2} \and
Baldo Faieta\inst{2} \and
Alex Filipkowski\inst{2} \and
Zhe Lin\inst{2} \and
John Collomosse\inst{1,2}
}%
\authorrunning{D. Ruta et al.}
%
\institute{University of Surrey \and
Adobe Research  \and University of Northumbria \and University of Edinburgh
}
\maketitle

\begin{abstract}

\noindent We present StyleBabel, a unique open access dataset of natural language captions and free-form tags describing the artistic style of over 135K digital artworks, collected via a novel participatory method from experts studying at specialist art and design schools. StyleBabel was collected via an iterative method, inspired by `Grounded Theory': a qualitative approach that enables annotation while co-evolving a shared language for fine-grained artistic style attribute description. We demonstrate several downstream tasks for StyleBabel, adapting the recent ALADIN architecture for fine-grained style similarity, to train cross-modal embeddings for: 1) free-form tag generation; 2) natural language description of artistic style; 3) fine-grained text search of style. To do so, we extend ALADIN with recent advances in Visual Transformer (ViT) and cross-modal representation learning, achieving a state of the art accuracy in fine-grained style retrieval. 


\end{abstract}

\section{Introduction}


             

\noindent Artistic style is the distinctive appearance of an artwork; i.e. how an artist has depicted their subject matter \cite{styledef}. Describing the artistic style of digital artwork is an open challenge for computer vision, which has focused on stylization, classification, and search in the style domain. However, automated style description has potential applications in summarization, analytics, and accessibility. For the first time, this paper shows that a set of descriptive tags, or even complete caption sentences, may be automatically generated to describe the fine-grained artistic style of an image -- distinct from its content \cite{oscar,meshedmemory,vinvl}, or the emotions it evokes \cite{artemis}. Our core contribution enabling this is \textbf{StyleBabel}, a novel dataset of fine-grained \cite{finegrained_cvpr11,finegrained_ieee17,finegrained_cvpr20} annotations describing the artistic style of $\sim$135K digital artworks\footnote{The dataset will be released for open access (CC-BY 4.0).}, collected from expert participants via a novel participatory method that forms a further contribution of this paper. Specifically, our novel contributions are: 

\begin{figure}
  \centering
  \includegraphics[width=\textwidth]{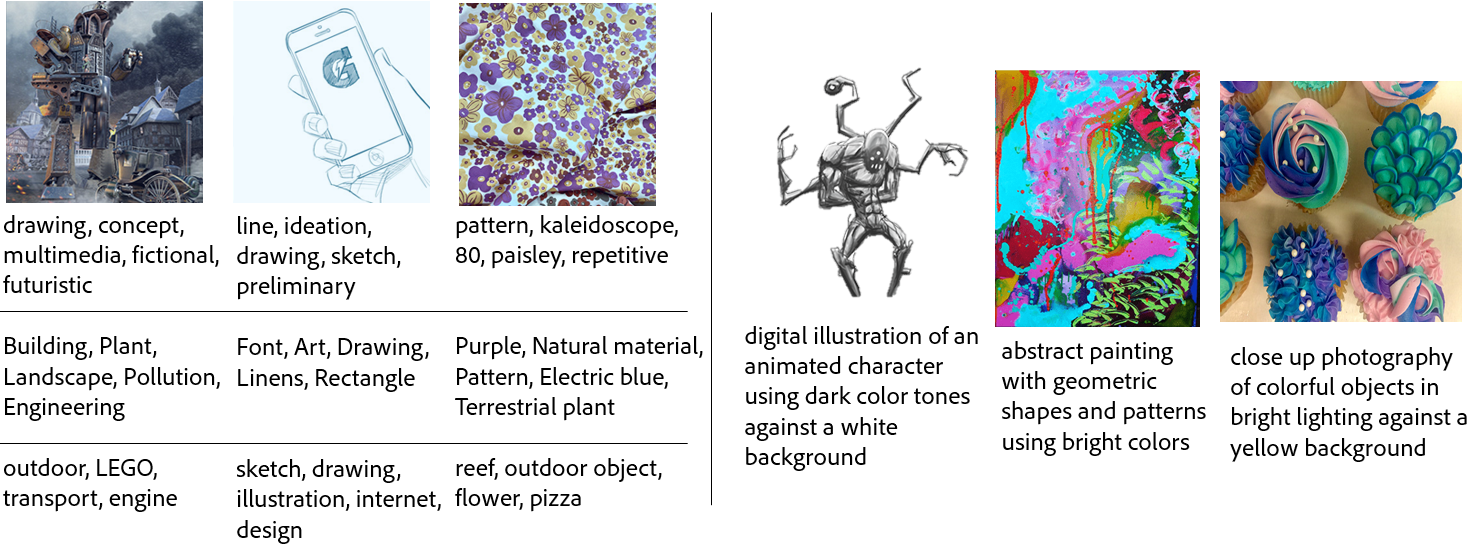}
    \caption{Results {\em generated} from models trained on StyleBabel to describe style using free-form tags (left) and captions (right). The top row of tags contains tags generated by our models trained with StyleBabel. The middle row contains tags from a commercial image tagging Google product, and the bottom from a similar commercial product from Microsoft. We show the importance of the style domain-specific information within StyleBabel, for generating relevant style captions.}
  \label{fig:teaser}
  \squeezeup
  \squeezeup
\end{figure}

{\bf 1. StyleBabel Dataset.}  
We present a new dataset of over $\sim$135K \{image, tags, natural language (NL) caption\} pairs for digital artwork images annotated by a combination of crowd-sourcing and 48 domain experts drawn from design and art schools. 
StyleBabel is the first dataset containing all three data types for every data sample. Furthermore, the images contained have vastly greater style variation than existing datasets \cite{wikiart,semart} which predominantly focus on small subsets of fine art, sometimes further limited to only European or Asian \cite{cvpr_rebuttal_b1,cvpr_rebuttal_r3_1,cvpr_rebuttal_r3_2}. Not only is StyleBabel's domain more diverse, but our annotations also differ. StyleBabel focuses on the visual appearance of images (which can include stroke/colouring type, lighting, shading, patterns, shapes, composition, medium, layout, theme etc.), we avoid external, high level information such as artists, time periods, surrounding meaning, emotions evoked, provenance facts, context or content. This enables new avenues for research not possible before, some of which we explore in this paper.

We do not seek to align our work to a formal ontology or definition of style (something heavily debated even by academics \cite{styledef,cvpr_rebuttal_r1_style_definition}). Instead, we explicitly build, evolve, and rely on the emergent structure of style information from the collective, harmonized experience of expert collaborators from art and design universities during our data collection process. In previous work \cite{nst_paper}, the working definition of style has been the similarity of gram matrices at specific layers in CNN backbones such as VGG \cite{vgg}. During our data pre-processing/initialization, ahead of style annotation, we similarly use a working definition as similarity in the ALADIN \cite{aladin} style model's embedding space, previously shown to accurately represent a variety of artistic styles in a metric space.  


During annotation, images are grouped into $\sim$6K moodboards (grids of style-consistent images). Each moodboard is annotated by a group of experts, both with tags and free-form captions, to yield a description of visual style. The vocabulary used for both annotation forms is unconstrained. The moodboard annotations are cross-validated as part of the collection process and refined further via the crowd to obtain individual, image-level fine-grained annotations.

{\bf 2. Grounded Annotation Methodology.} We present a new data annotation methodology inspired by {\em Grounded Theory} (GT)~\cite{ground,charmaz}, a qualitative research method used in the humanities and social sciences.   In GT, participant groups engage in an unconstrained data clustering exercise while simultaneously evolving a shared vocabulary for describing those clusters. Working with these disciplines, we adapted this process to evolve a shared vocabulary for annotation while annotating groups (`moodboards') of images with similar artistic styles. The moodboards are obtained by clustering artworks within the ALADIN fine-grained embedding for style similarity \cite{aladin}. Our iterative methodology comprises individual and participatory group stages and a validation stage.

{\bf 3. Artistic Retrieval and Description.}  We incorporate a Visual Transformer~\cite{visualtransformer1} into a fine-grained visual style representation architecture~\cite{aladin}, (ALADIN-ViT). ALADIN-ViT provides state of the art performance at fine-grained style similarity search. We train models for several cross-modal tasks using ALADIN-ViT and StyleBabel annotations. Using CLIP~\cite{clip}, we train with StyleBabel to generate free-form tags describing the artistic style, generalizing to unseen styles. We show that we may apply these tags for text based style search. We similarly demonstrate the synthesis of descriptive natural language captions for digital art.


\section{Related Work}

\noindent Representation learning for visual style has focused primarily on neural style transfer (NST) and style classification. 

{\bf Style Transfer.}  Classical approaches learned patch-based representation of style by analogy from paired data.  Gatys et al. ~\cite{nst_paper} enabled NST by extracting disentangled representations separating content and style from unpaired data, using a Grammian computed across layers of a pre-trained VGG-19 \cite{vgg} model. Similarly, feed-forward networks used the Grammian to train fast encoder-decoder models for NST specialized to pre-trained styles \cite{Ulyanov2016,Johnson2016}.  Extensions to multi-scale \cite{Wang2017} and video \cite{Ruder2016} NSTS were later presented. To relax the constraint to pre-trained styles, feature based NST was proposed using Adaptive Instance Normalization (AdaIN) \cite{adain_paper,google_adain}.  This approach was further generalized by the whitening and coloring transform (WCT), which matched feature covariances \cite{Li2017}.  Recently unsupervised style transfer was enabled via MUNIT \cite{munit} and swapping autoencoder \cite{Park2020}.  The latent style codes of encoder-decoder networks for NST were recently adapted for style-based visual representation learning, using weak supervision to learn a metric embedding for fine-grained similarity using the ALADIN model \cite{aladin}.

{\bf Style analytics} via classification has been explored for digital artwork \cite{bam}, smaller fine-art collections  \cite{Zujovic2009,Karayev2014}, product designs \cite{Bell2015}, and to identify both painters \cite{Cetinic2013} and genres \cite{Shamir2010}.  Style-based visual search has also been proposed for coarse-grained style using triplet  \cite{Collomosse2017} and constrastive training for fine-grained style via ALADIN \cite{aladin}.  

{\bf Style datasets.}  No prior dataset of fine-grained artistic style description exists.  However, several annotated datasets of artwork have been produced. {\bf Behance Artistic Media -- BAM} \cite{bam} comprises 2M diverse digital artworks from the \texttt{Behance.net} platform, with 7 coarse-grained style and 4 emotion tags and no descriptions.    {\bf Omniart} (432K images) and Wikiart (81K) are datasets of fine art with associated metadata but no descriptions or tags.  The {\bf SemArt} dataset \cite{semart} focuses on very high level contextual semantic information, rarely containing style (visual appearance) information, and narrowly focuses solely on 8th-19th century European fine art. The {\bf BAM-FG} (BAM Fine-grained) dataset comprises 2.62M images grouped into 310K style-consistent groupings but no descriptive text or labels.   Our work also uses \texttt{Behance.net}, to provide expert style tags and natural language descriptions over 135K images.
The \textbf{AVA} dataset \cite{ava_dataset} studies aesthetics information in \textit{photographic} images only, and the follow-up \textbf{AVA-captions} dataset \cite{ava_captions} adds captions for these, sourced from noisy internet comments sections.
Recently, ArtEmis~\cite{artemis} released non-expert annotations capturing the emotions felt by viewers of fine art in WikiArt. Our proposed StyleBabel dataset is aligned to this contemporary work in that it also seeks to ascribe text to visual art. However, our focus is also on digital, not just fine artwork. We also differ in that our annotation is led by expert students in specialized art and design schools, not exclusively by non-expert crowd-workers.  Notably, our annotations focus on the style alone, deliberately {\em avoiding} the description of the subject matter or the emotions that matter evokes.  {\bf ArtEmis} instead describes {\em exclusively the emotions evoked} by both the style and content of the artwork, both of which feature in the descriptions. To recap, StyleBabel is unique in providing tags and textual descriptions of the artistic style, doing so at a large scale and for a wider variety of styles than existing datasets, with labels sourced from a large, diverse group of experts across multiple areas of art.

{\bf Image captioning} and visual question answering  methods \cite{lstm_caption} initially learned LSTM language models, leveraging semantic image embeddings e.g. via ResNet/ImageNet. Later image captioning work \cite{oscar,meshedmemory,vinvl,unifiedVL} made use of object detection; regions of interest (ROI). Their relationships yielded improved semantics captioning models, though often due to the bias of co-present context that hinted at the image narrative. This is incompatible with our domain of artistic style, where this localization bias is not something we can use. Recently, there has been an influx of research combining the visual and text domains \cite{clip,dalle,ramesh2021zeroshot}. Established methods \cite{templatepaper1,templatepaper2},  have primarily functioned by generating captions from templates. Though this approach benefits from generating grammatically correct captions, its inflexibility has led to diminished interest in its application, despite recent implementation by OpenAI’s CLIP \cite{clip}. Here, captioning capabilities generate tags and insert them into templates, using two encoders -- for text and image modalities. This model then performs cross-modal training via contrastive loss.
More recently, Attention-on-attention \cite{aoa} can generate state-of-the-art quality captions from an image embedding by filtering out irrelevant attention results.
VirTex \cite{virtex} recently demonstrates that caption annotations are more efficient for representation learning of images, with better or comparable representation quality on ImageNet despite much less training data.

{\bf Grounded Theory (GT)} \cite{ground} is a qualitative method used in the humanities and social sciences to codify data --  i.e. to identify and name apparent or emergent patterns across different data sources and types. Through interpretation, participants collaboratively agree on an initial set of summative ‘open’ codes that are then iteratively combined into larger groups, until eventually, participants arrive, by consensus, at the common themes in the data \cite{charmaz}.  We adopt this approach to collecting expert annotation to describe the artistic style of clusters of digital artwork. Free-form textual input from various participants can vary in writing style, creating a very noisy dataset. GT mitigates this by guiding participants to align their responses to a consistent format.

\section{StyleBabel Dataset}

\noindent StyleBabel is a new dataset for cross-modal representation learning.  It comprises 135k digital artwork images from the public creative portfolio website \texttt{Behance.net} (in turn, available via the BAM dataset).  Each image is annotated with a set of keyword tags and natural language descriptions {\em 'captions'} describing its {\em fine-grained artistic style} -- the distinctive appearance of the image  -- in the English language.  We focus mainly on attributes that we can visually depict, rather than more high level and abstract concepts such as emotions \cite{artemis}.   
StyleBabel enables the training of models for style retrieval and generates a textual description of fine-grained style within an image: automated natural language style description and tagging (e.g. style2text).  
We train state of the art proof of concept models for these tasks using our dataset in Sec.~\ref{sec:exps}.

\subsection{Study Context}

\noindent We determined via early initial trials with crowd annotation platforms (AMT) that the quality, coarseness, and diversity of data generated by non-experts is inadequate for style description tasks. We collaborated with graduate schools specializing in digital art and design to address this. Together with academic experts at these schools, we designed a novel multi-staged participatory method to enable novel style vocabulary gathering, tagging, and caption generation, recruiting 48 expert staff and student participants. We particularly sought (but did not make a prerequisite) participants familiar with Behance. The final cohort comprised a representative balance of gender and ethnic background of both graduate and undergraduate cohorts - anecdotally, the gender split was equal. Expertise was at that of a final year undergraduate student, with more senior faculty members participating in the discussions. Their program's specialisms were primarily communication design (graphics, illustration), industrial design, fashion, and animation.  

\begin{figure*} [t!]
    \begin{center}
            \includegraphics[width=0.32\linewidth,height=2.5cm]{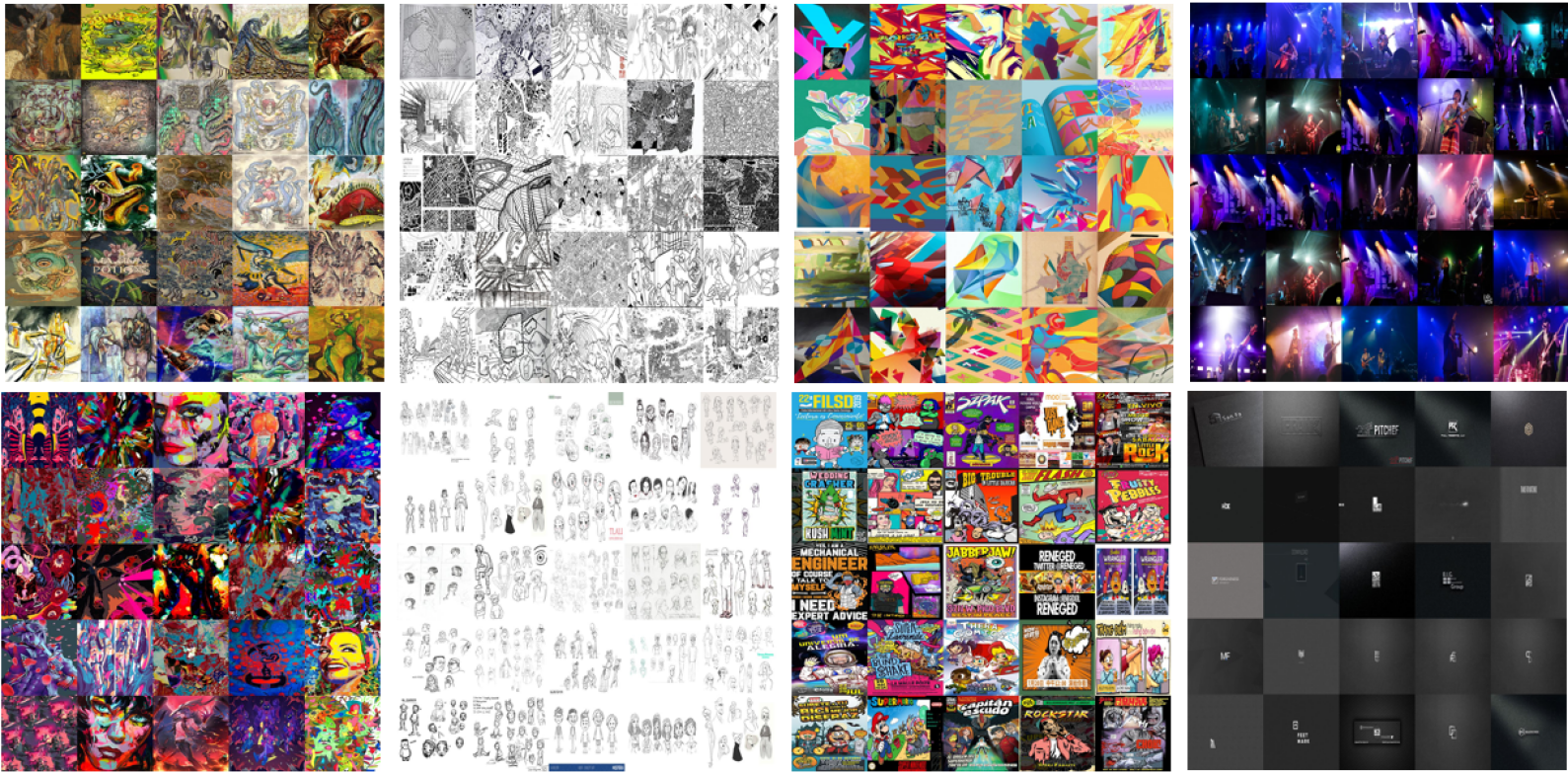}  ~~
        \includegraphics[width=0.65\linewidth,height=2.5cm]{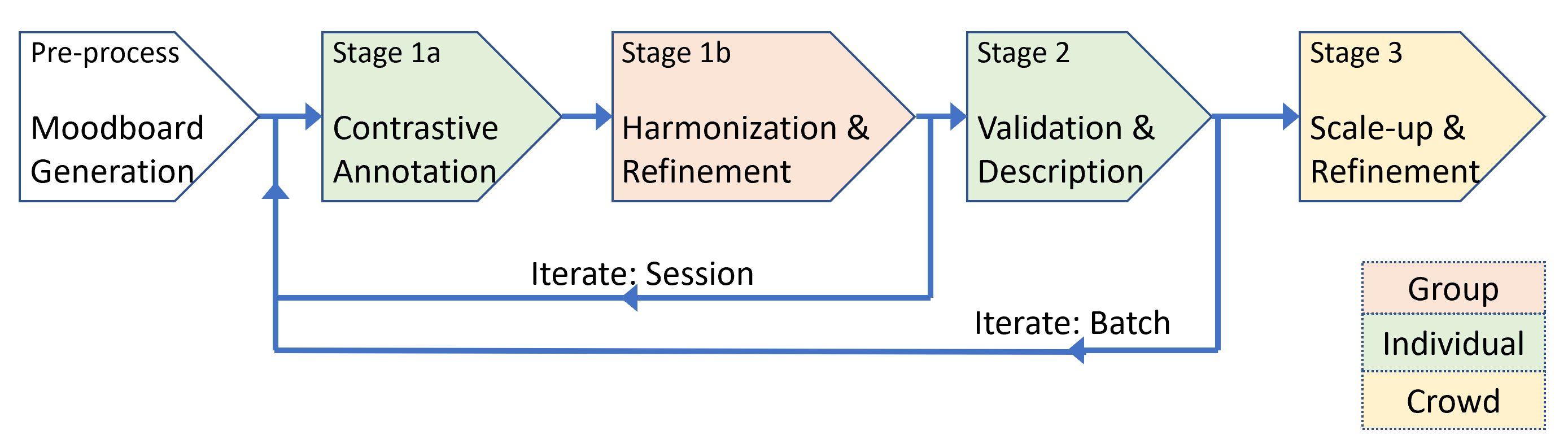}  
    \end{center}
    \caption{The StyleBabel Grounded Annotation process.  Moodboards (examples, left) are annotated via an iterative process (right) that encourages groups of participants to evolve and converge to a shared language for describing style.  See subsec.~\ref{sec:process} for stage descriptions.}
    \label{fig:workflow}
    \squeezeup
    \squeezeup
\end{figure*}

We executed the group exercise online using a \textit{collaborative whiteboard} interactive platform\footnote{https://miro.com/},  that provides capabilities for multiple users to move and annotate components freely in real time. We built 1300 unique pages of moodboards (Fig.\ref{fig:workflow}) to allow participants to move sticky notes with related tags naturally in a collaborative process. The process simulated in-person group collaboration, enabling the formation of tag clusters and aligning to processes of GT codification. The combined use of Miro and Zoom supported real-time spatial organization of information and associated discussion. Workers were paid significantly higher than the national minimum, with a total dataset cost of approximately \$160k, which we freely contribute as CC-BY 4.0.

\subsection{StyleBabel Grounded Annotation}
\noindent StyleBabel {\em does not} aim to develop an ontology to categorize style, with agreement on a diverse ontology eluding art practitioners \cite{styledef}. Yet, consistency of language is essential for learning of effective representations. Therefore we propose an annotation methodology that enables annotations at scale (multiple participants) and encourages co-evolution of a harmonized natural language to describe the style.

Our annotation process instead is inspired by Grounded Theory (GT)~\cite{ground,charmaz}; a qualitative method often used for data analysis in the humanities and social sciences.  A systematic research process to `codify' empirical data, identify themes from the data, and associate data with those themes.  This process is distinct from fitting (or `annotating') data to pre-existing categories. GT is an iterative process in which participants co-evolve a language to describe the data as they work on clustering and labeling it with that shared language.  Concretely, GT often begins with a discussion around a subset of the data during which clusters are formed.  Data is moved freely between clusters during the debate, from which a shared understanding and, ultimately, a shared terminology evolves for describing those clusters. With further data, this language identifies and names patterns apparent or emergent in the data.

\subsubsection{Grounded Annotation Process}
\label{sec:process}

We propose a multi-stage process for compiling the StyleBabel dataset comprised of initial individual and subsequent group sessions and a final individual stage. Each batch of these sessions was estimated to take around 10 hours, run over a week, and run four sets over four weeks.  

Experts annotate images in small clusters (referred to as image `moodboards').  Moodboards are obtained by automatically clustering artworks within a fine-grained style embedding~\cite{aladin}. Our annotation process thus pre-determines the clusters for expert annotation. Still, it encourages expert groups to evolve a harmonized language during the iterative annotation process (as in GT) to improve data consistency. We refer to this process as `grounded annotation'.

\vspace{1ex} 
\noindent \textbf{Pre-processing} - \textit{Moodboard generation (Automated)}
\vspace{1ex}

We downloaded \noindent an initial dataset of 150 million digital artwork (static image assets) from Behance.net. We encode all images into a metric search embedding~\cite{aladin}, that we then clustered into 6.5K clusters, with $L_2$ distance to identify the 25 nearest image neighbors to each cluster center. The images from each cluster were arranged in a $5  \times 5$ grid for presentation as a moodboard. Thus, we start the annotation process using 6,500 moodboards (162.5K images) of 6,500 different fine-grained styles.\footnote{We redacted a minimal number of adult-themed images due to ethical considerations}. The extremely high data density from this internet-scale data corpus ensures that the small clusters formed are very stylistically consistent.
Fig. \ref{fig:workflow} (left) shows examples of moodboards. 

\vspace{1ex} 
\noindent \textbf{Stage 1a} - \textit{Contrastive Annotation (Individual)}
\vspace{1ex}    
    
\noindent Participants were individually presented with a pair of 5x5 moodboards and asked to generate a list of textual tags (`{\em style attributes}') that occur in one moodboard, but not another, and a list of style attributes which are shared by both.  The moodboards were sampled such that they were close neighbors within the ALADIN style embedding. In the annotation, comparative language (e.g. `X is brighter than Y') was not permitted to encourage standalone descriptions (e.g. `bright' was allowed). 
This paired approach encouraged the suggestion of fine-grained style attributes, supporting participants to suggest characteristics that may otherwise not have been considered when looking at individual styles. Multiple participants annotated each moodboard in this way to produce a rich initial set of attributes.

\begin{figure}[t!]
  \centering
  \small
      \centering
      \begin{adjustbox}{width=1.05\linewidth}
          \begin{tabular}{p{0.16\linewidth}|p{0.40\linewidth}p{0.40\linewidth} }

            & \vspace{-0.2cm} \raisebox{-\totalheight}{\includegraphics[width=0.20\textwidth]{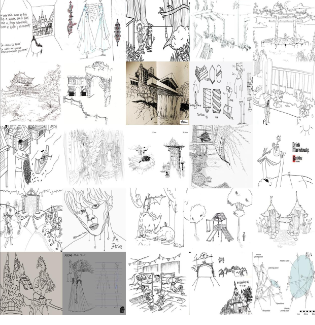}} & \vspace{-0.2cm}  \raisebox{-\totalheight}{\includegraphics[width=0.20\textwidth]{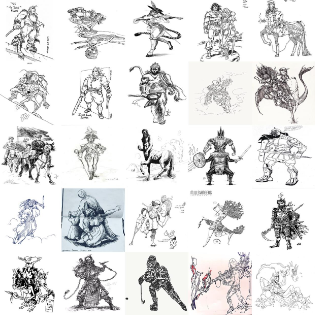}} \\ 
            
            \hline
            Changed (\textcolor{fig_red}{Removed} or \textcolor{fig_green}{added} tags in Stage 1b) & \textcolor{fig_red}{line, both have the same white backgrounds with very little tonal variations/ just block colours, no difference in tones, bright white background, b+w, constructed by lines, no colour, hand drawing} \textcolor{fig_green}{linear, pen and ink}   & \textcolor{fig_red}{b+w} \textcolor{fig_green}{black and white, pen and ink, fantasy, intricate, white space, central composition, linear, illustration} \\
            \hline
            Final tags after Stage 1b cleaning & ink work, sketches, bright, \underline{black and white}, white background, clean, monotonal, \underline{drawing}, simple, \underline{illustration}, \underline{linear}, \underline{pen and ink} & graphic, \underline{drawing}, \underline{black and white}, \underline{pen and ink}, fantasy, intricate, white space, central composition, \underline{linear}, \underline{illustration} \\
            
          \end{tabular}
      \end{adjustbox}
    \vspace{0.1cm}
    \caption{Before and after harmonization (Stage 1b), showing the benefits of the GT approach. Two moodboards receive linguistically different tag sets in Stage 1a (individual) but are conformed to the shared vocabulary evolved by the participant group in Stage 1b (group). The moodboard styles and the tag sets differ but draw upon a shared vocabulary (underscored tags) despite the board not being shown in the same session to workers. Tags removed are in red, and additions are in green.}
    \label{fig:ex1_lang_harmonisation}
    \squeezeup
    \squeezeup
\end{figure}

\vspace{1ex}
\noindent \textbf{Stage 1b} - \textit{Harmonization and Refinement (Group)}
\vspace{1ex}    
    
\noindent A collaborative workspace was synthesized within Miro, in which 5 moodboards and their associated style tags from Stage 1a are displayed (as `sticky notes' below each moodboard). 
    
\begin{figure}[t!]
        \begin{center}
            \includegraphics[width=\linewidth,height=4cm]{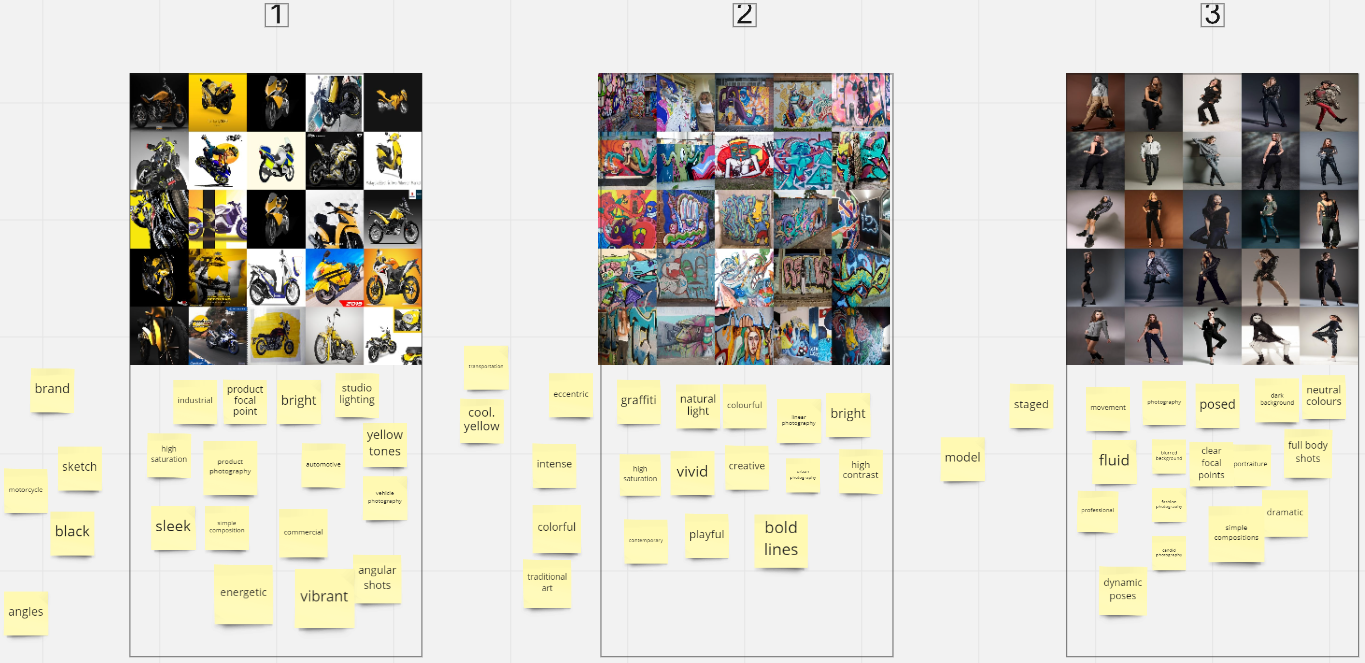}  \\
        \end{center}
        \squeezeup
        \caption{Virtual environment for collaborative Stage 1b. Style attributes (as `sticky notes') are  added, modified and removed from each moodboard to harmonize and filter language.}
        \label{fig:p2}
        \squeezeup
        \squeezeup
    \end{figure}

After an initial briefing and group discussion, each group considered moodboards collectively, one moodboard at a time. All participants were asked to add new tags to the pre-populated list of tags that we had already gathered from Stage 1a (the individual task), modify the language used, or remove any tags they agreed were not appropriate. Each moodboard was considered `finished’ when no more changes to the tags list could be readily determined (generally within 1 minute). Workers spent at most 2h 15m per session, including all breaks. At least one facilitator was always present throughout to ensure high engagement. Fig. \ref{fig:p2} displays an example of moodboards presented during this part of the study via the Miro platform.  
Fig.~\ref{fig:ex1_lang_harmonisation} provides an illustrative example of language harmonization. Two Stage 1a moodboards with initially very different language but similar style resulted in similar attributes post-Stage 1b. 
    
    
\noindent \textbf{Stage 2} - \textit{Validation and Description (Individual)}
\vspace{1ex}

\noindent The participants completed the final stage individually. Each participant was presented with a random collection of 5 moodboards and a set of tags generated during the previous sessions for just one of these moodboards. Participants engaged in ESP-like game \cite{Anh2004} to identify which moodboard the tags had been generated to describe, to verify accuracy. Further, we then asked them to create natural language captions, using as many presented tags as possible. We stressed to include as many tags as possible as by now, they had all been thoroughly cleaned and refined.


\noindent \textbf{Stage 3} - \textit{Scale-up and Refinement  (Individual)}
\vspace{1ex}

\noindent Following these sessions with our subject experts, we ran a different task with non-experts (workers on Amazon Mechanical Turk (AMT)). We asked at an \textit{image level} (rather than at the 5x5 cluster/moodboard level as previously) to discard any tags they considered not appropriate. Though the moodboards presented to these non-expert participants are style-coherent, there was still variation in the images, meaning that certain tags apply to most but not all of the images depicted. This step helped us to refine the final tags to individual images further. Unlike novel vocabulary generation, this verification step can be readily performed by non-experts, as detecting invalid tags is a much easier task than novel tag generation. 
We collected 3 responses per image and performed majority consensus to determine the final tags to use at the per-image level. 

     
We additionally performed a large-scale crowd annotation exercise to individualize cluster-level captions to individual images. Trained workers were presented with individual images, its tags, and the moodboard caption and were asked to compose (potentially many)  natural language captions using the tags and caption,  ensuring the full set of tags were incorporated across those sentences.
A constant set of workers were trained with feedback, for several months, together with a Quality Control (QC) process to ensure high quality annotation. The QC process included grammatical correctness checking, and rejection of any description of content or emotion the descriptive captions.

\subsubsection{Language processing}

Aside from the crowd  data filtering, we cleaned the tags emerging from Stage 1b through several steps, including removing duplicates, filtering out invalid data or tags with more than 3 words, singularization, lemmatization, and manual spell checking for every tag. The spell checking step was carried out in 3 passes. The accuracy of the validation step in Stage 2 was found to be 90\%.

\begin{figure}[h]
    \begin{center}
        \includegraphics[width=0.4\linewidth]{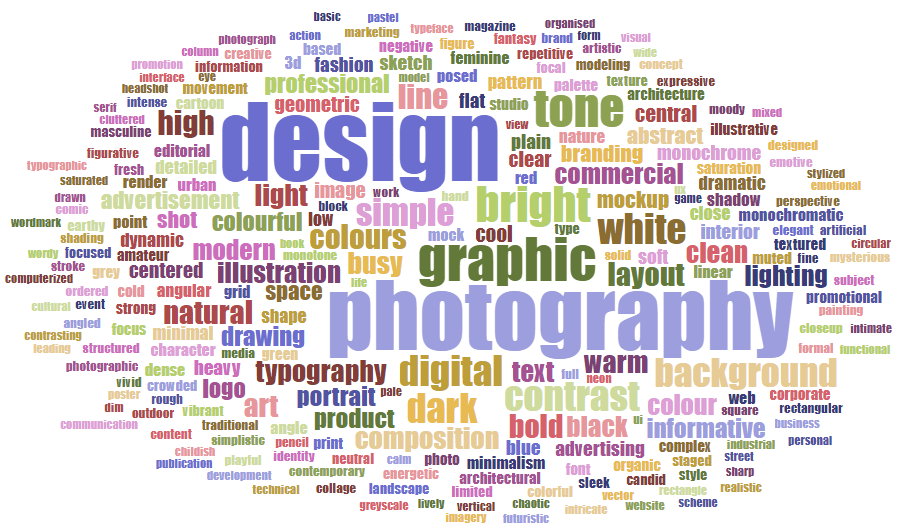}  \\
    \end{center}
    \squeezeup
    \caption{Visualizing the top 250 tags captured within the StyleBabel annotation over 135K images.}
    \squeezeup
    \label{fig:wordcloud}
    \squeezeup
\end{figure}

The final StyleBabel dataset contains 135k images with an average of 12.8 tags per image, over 6k style groups (of the 6,500 initially sampled, with 6k completed by workers in the available time).  The tags dictionary contains 3,151 unique tags, and the captions contain 5,475 unique words. Prepositions, determiners, and conjunctions were filtered out. Cardinals (eg. \textit{80s}) are kept. 9 cardinals, 2098 nouns, 500 verbs, 55 adverbs, and 482 adjectives are captured. Fig. \ref{fig:wordcloud} visualizes a word cloud from the 250 most common style attributes in StyleBabel, and Tbl.~\ref{tab:data_examples} shows the richness of the tags and captions.

\begin{table*}
  \centering
  \small
      \centering
      \begin{adjustbox}{width=\textwidth}
          \begin{tabular}{p{0.1\linewidth}|p{0.29\linewidth}|p{0.25\linewidth}|p{0.23\linewidth}|p{0.26\linewidth}}
          Image & 
          \vspace{-0.2cm} \raisebox{-\totalheight}{\includegraphics[width=0.18\textwidth]{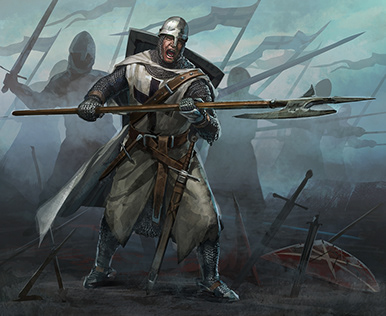}} & 
          \vspace{-0.2cm} \raisebox{-\totalheight}{\includegraphics[width=0.15\textwidth]{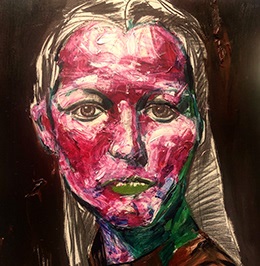}} & 
          \begin{center}
            \vspace{-0.55cm} \raisebox{-\totalheight}{\includegraphics[width=0.1\textwidth]{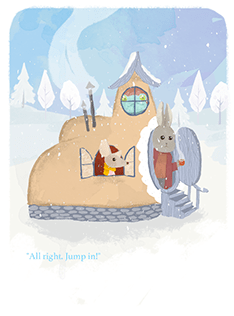}}    
          \end{center} & 
          \vspace{-0.2cm} \raisebox{-\totalheight}{\includegraphics[width=0.19\textwidth]{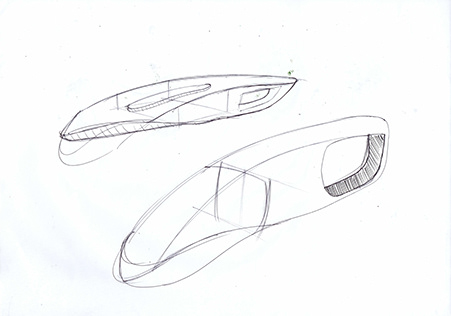}} \\
          
          \hline
          Tags &
          dim, concept, action, fantasy, powerful, digital, photography, animated, prototype, masculine, detailed, professional, lighting & 
          abstract, moody, portrait, oil, painting, drawing, artistic, melancholic, pleasing & 
          cold, digital, book, bright, colors, drawing, child, stroke, busy, clear, illustration, festive, blue & 
          experimental, analog, line, development, black, drawing, sketch, figure, commercial, white, scamp, stroke, product, pencil, rough, thin, isometric  \\
          
          \hline
          Caption & 
          Fantasy themed digital illustration featuring an animated male character, dim highlighting and a hazy, dark and cluttered background. The illustration highlights the powerful masculine character with sharp objects around. & 
          Portrait oil painting of a female character featuring abstract shapes and psychedelic patterns against a dark background. The artistic artwork is melancholic and using thin repetitive strokes and shades. & 
          Digital bright fantasy anthropomorphism cartoon illustration created with soft diffused blended hues, brush strokes, lines, and geometric forms in neutral and cool tones. & 
          Analog experimental sketches with thin pencil strokes and lines. The isometric drawing expresses commercial product development. \\
           
          \end{tabular}
      \end{adjustbox}
    \caption{Excerpt of StyleBabel dataset. Four images and the corresponding tags and captions collected via our Grounded Annotation.}
    \label{tab:data_examples}
\end{table*}

\section{Visual Embedding (ALADIN-ViT)}
\label{sec:aladin}

\begin{figure}
    \begin{center}
        \includegraphics[width=0.7\linewidth]{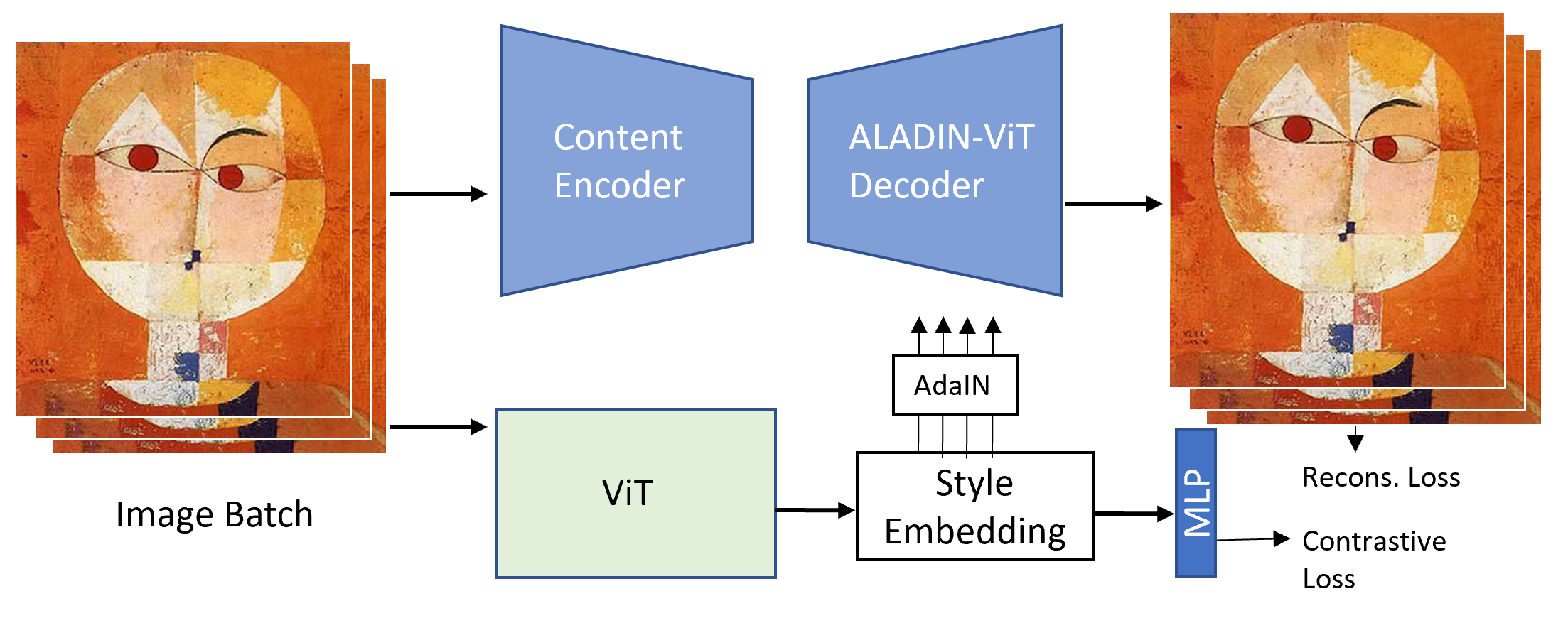}  \\
    \end{center}
    \squeezeup
    \squeezeup
    \caption{ViT-ALADIN architecture used for StyleBabel experiments; as per ALADIN \cite{aladin} but swapping the style encoder for ViT \cite{vit_paper} (change in green) and retrained end-to-end on BAM-FG.}
    \label{fig:avit}
    \squeezeup
\end{figure}

\noindent ALADIN is a two branch encoder-decoder network that seeks to disentangle image content and style. It works by pooling Adaptive Instance Normalization (AdaIN) statistics across multiple layers in the style encoder to produce an embedding for fine-grained style similarity using a VGG-19 convolutional backbone for the style encoder. 
In our later experiments, we require to use a Visual Transformer \cite{vit_paper} (ViT) model for the vision domain. To achieve this, we adapt ALADIN to use ViT as the style encoder backbone; we refer to this as ALADIN-ViT (Fig. \ref{fig:avit}). We retrain ALADIN-ViT on BAM-FG following the same training method ~\cite{aladin},  i.e. using both the reconstruction loss and the weakly supervised contrastive loss and using the implicit style grouping in BAM-FG. Having swapped the style encoder for a transformer, it is no longer possible to sample AdaIN statistics from feature maps in the encoder. However, we retain the use of the style code as pairs of values to be split and used in the decoder stage, again keeping the size of the style code at double the number of feature maps in the decoder. 

We achieve the state of the art fine grained style retrieval accuracy on the BAM-FG test partition (Tbl. \ref{tab:avit_aladin}) at \textbf{64.48} Top-1, beating not only ALADIN (58.98) but also their fused variant (62.18), which incorporates ResNet embeddings into a concatenated embedding. We, in part, attribute the gains in accuracy to the larger receptive input size (in the pixel space) of earlier layers in the Transformer model, compared to early layers in CNNs. Given that style is a global attribute of an image, this greatly benefits our domain as more weights are trained on more global information.
\begin{table}[htbp]
  \centering
  \small
      \centering
      \begin{tabular}{ll|r}
        \hline
        Data & Model &  IR-1 \\
        \hline
        BAM-FG-C$_5$ & ALADIN & 58.98  \\ 
        BAM-FG-C$_5$ & ResNet & 45.22  \\
        BAM-FG-C$_5$ & ALADIN (Fused) & 62.18 \\
        \hline
        BAM-FG-C$_5$ & ViT & 57.91 \\
        BAM-FG-C$_5$ & ALADIN-ViT & \textbf{64.48} \\
        \hline
      \end{tabular}

    \caption{Fine grained style retrieval on the BAM-FG dataset~\cite{aladin} of proposed method ALADIN-ViT, compared to previous methods}
    \label{tab:avit_aladin}
  \squeezeup
  \squeezeup
  \end{table}

\section{StyleBabel Experimental Setup}
\label{sec:exps}
{\bf Data Partitions.} We define train/validation/test partitions within StyleBabel for our experiments as follows. Splits were separated on a moodboard basis, to avoid overlap. The training split has 133k images in 5,974 groups with 3,167 unique tags at an average of 13.05 tags per image. The validation and test splits contain 1k unique images for each validation and test, with 1,256/1,570/10.86 and 1,263/1,636/10.96 unique tags/groups/average tags per image. Captions have an average of 2.4 sentences, with an average of 19.3 words, from 6119 unique words. 1000 samples were extracted for each of the test/validation splits. 

{\bf Implementation.} The models were trained with a maximum batch size of 11k on a 12GB GTX Titan, with a learning rate of 0.003, Adam optimizer, and weight decay of 1e-6. Logit accumulation \cite{aladin} was employed to reach the maximum batch size possible in the GPU VRAM capacity. We trained all models to convergence. The learning rate was decayed using cosine annealing, as per SimCLR \cite{simclr}. For the VirTex captioning experiments, a batch size of 105 was used, on a V100, with a learning rate of 2e-4.

\begin{figure*}
  \centering
  \small
      \centering
      \begin{adjustbox}{width=0.85\textwidth}
          \begin{tabular}{p{0.20\linewidth}p{0.20\linewidth}p{0.20\linewidth}p{0.20\linewidth}p{0.20\linewidth} }

            \vspace{-0.2cm} \raisebox{-\totalheight}{\includegraphics[width=0.20\textwidth]{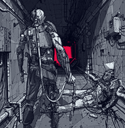}} & \vspace{-0.2cm} \raisebox{-\totalheight}{\includegraphics[width=0.20\textwidth]{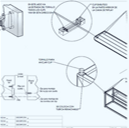}} & \vspace{-0.2cm} \raisebox{-\totalheight}{\includegraphics[width=0.20\textwidth]{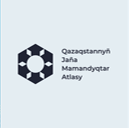}} & \vspace{-0.2cm} \raisebox{-\totalheight}{\includegraphics[width=0.20\textwidth]{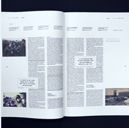}} & \vspace{-0.2cm} \raisebox{-\totalheight}{\includegraphics[width=0.20\textwidth]{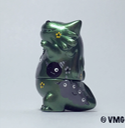}} \\ 
            fictional, anime, imaginary, expressionist, manga & 
            design, technical, planning, scamp, calculated & 
            development, ligature, minimal, symbolic, logo & 
            column, academic, article, spread, informative & 
            3d, sculptural, product, creativity, miniature \\
            
            \vspace{-0.2cm} \raisebox{-\totalheight}{\includegraphics[width=0.20\textwidth]{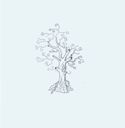}} & \vspace{-0.2cm} \raisebox{-\totalheight}{\includegraphics[width=0.20\textwidth]{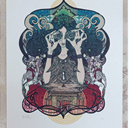}} & \vspace{-0.2cm} \raisebox{-\totalheight}{\includegraphics[width=0.20\textwidth]{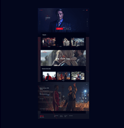}} & \vspace{-0.2cm} \raisebox{-\totalheight}{\includegraphics[width=0.20\textwidth]{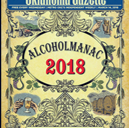}} & \vspace{-0.2cm} \raisebox{-\totalheight}{\includegraphics[width=0.20\textwidth]{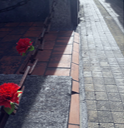}} \\ 
            
            line, thin, drawn, ink, pen & expressive, personal, artistic, creative, whim & 
            web, interface, format, media, ui & 
            chaotic, mural, complex, decay, overload & 
            angular, geometric, modernist, linear, contemporary \\
            
          \end{tabular}
      \end{adjustbox}
    \caption{Style2Text tag generation experiment, using CLIP trained with ALADIN-ViT encoder.  Top 5 tags shown for each image.}
    \label{fig:tags_viz}
\end{figure*}



\section{Experiments and Discussion}

\noindent We illustrate the potential of our StyleBabel dataset for three cross-modal learning tasks:

\noindent {\bf 1. Style Auto-tagging:} (style2text) Using StyleBabel tags, we train a CLIP~\cite{clip} model to learn a cross-modal embedding between image and text embeddings.  We generate several tags for unseen StyleBabel images and explore the generated tags' accuracy and the model's ability to generalize.

\noindent {\bf 2. Style Description:} (style2text) We similarly make use of the natural language captions collected in StyleBabel, and showcase the generation of natural language captions describing the style of images. 

\noindent {\bf 3. Text Based Style Retrieval:} (text2style) We explore the efficacy of our tag generation for text based style search.

\subsection{Style Auto-tagging (style2text)}
\label{sec:clip_tags}

\noindent Recent literature in image captioning has transitioned to making use of object detectors in their model pipelines. This makes sense in semantics, as such features are most often localized to a subset of the image. Style, however, is typically a global attribute of an image, and object detectors are not compatible. Style is more abstract and seldom localized to any specific region of an image. We use the CLIP \cite{clip} training methodology to learn a joint embedding space between the publicly available CLIP text encoder and our new vision transformer (ALADIN-ViT). CLIP is traditionally formed of two transformers, the first for text encoding and the second for image encoding. Two MLP heads are trained together through contrastive loss to learn a joint text/image embedding, adding invariance to the modality. 
We freeze both pre-trained transformers and train the two MLP layers (ReLU separated fully connected layers) to project their embeddings to the shared space. We follow CLIP, employing contrastive loss to drive learning, with the same training set-up.


When using the model for inference, we pass the entire dictionary of available tags through the text encoder and multi-modal MLP head to generate text embeddings. Next, we infer the image embedding using the image encoder and multi-modal MLP head, and  calculate similarity logits/scores between the image and each of the text embeddings. We evaluate accuracy via WordNet similarity \cite{wordnet} using the \textit{nltk}\footnote{https://www.nltk.org/} library to first compute synsets for the $N$ ground truth tags for an image. Next, we sort the tags in the dictionary by the logit scores following the embedding inference similarity. We select the top $N$ "retrieved" tags, and for each, and calculate the WordNet similarity to each ground truth tag. The similarity ranges from 0 to 1, where 1 represents identical tags. We save the top value (the similarity score for the closest/most relevant ground truth tag) and repeat it for each test image. These values are averaged to form our \textit{WordNet score}. 
We use this instead of precision/recall, as multiple tags in the dataset can represent very similar (but not identical) concepts. A soft score better encapsulates the perceived accuracy of the tags.

We experiment with training CLIP on variants of StyleBabel, presenting results in Tbl.\ref{tab:clip_results}. In particular, we quantify the difference in quality between collecting annotation via non-expert crowd annotation (StyleBabel-mturk) and gathering expert annotations using our GA process (StyleBabel/ALADIN-ViT). We also show the value of the final stage, where we refine tags to the image-level (FG) rather than moodboard-level (coarse). We train the MLP heads atop the CLIP image encoder embeddings (the 'CLIP' model) and atop embeddings from our ALADIN-ViT model (the 'ALADIN-ViT' model). The former is not based on the ALADIN-ViT style embedding and underperforms by 40\%.  The best performing model is the proposed ALADIN-ViT trained via StyleBabel data collected using GA on FG labels, with a WordNet score of \textbf{0.352}, double the CLIP~\cite{clip} baseline. Fig. \ref{fig:tags_viz} shows some examples of tags generated for various images, using the ALADIN-ViT based model trained under the CLIP method with StyleBabel (FG).

\begin{table}[t!]
  \centering
  \small
      \centering
      \begin{tabular}{l|l|c}
        \hline
        Data & Model & WordNet score \\
        \hline
        CLIP Webscale & CLIP \cite{clip} baseline & 0.168 \\
        \hline
        StyleBabel-mturk & ALADIN-ViT & 0.164 \\
        \hline
        StyleBabel (coarse) & CLIP \cite{clip} & 0.187 \\
        StyleBabel (coarse) & ALADIN-ViT & 0.225 \\
        \hline
        StyleBabel (FG) & CLIP & 0.215 \\
        StyleBabel (FG) & ALADIN-ViT & \textbf{0.352} \\
        \hline

      \end{tabular}

    \caption{Ablation experiments for tag generation under the CLIP training setting. We show the benefit of annotating StyleBabel via the proposed GT annotation versus non-expert crowd-sourcing (StyleBabel-mturk). We further demonstrate the benefits of tag annotations individualised to the image-level (FG), compared to cluster-level (coarse).}
    
    \label{tab:clip_results}
   \squeezeup
   \squeezeup
\end{table}


\begin{figure*}
  \centering
  \small
      \centering
      \begin{adjustbox}{width=\textwidth}
          \begin{tabular}{p{0.11\linewidth}p{0.22\linewidth}p{0.25\linewidth}p{0.22\linewidth}p{0.22\linewidth}p{0.23\linewidth} }
            &
            \vspace{-0.2cm} \raisebox{-\totalheight}{\includegraphics[width=0.18\textwidth,height=0.15\textheight]{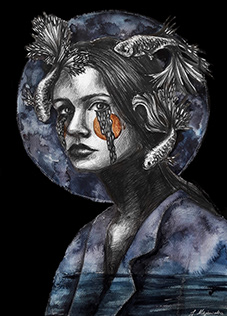}} & \vspace{0.4cm} \raisebox{-\totalheight}{\includegraphics[width=0.22\textwidth]{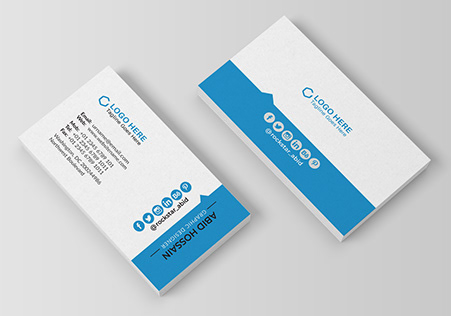}} & \vspace{-0.2cm} \raisebox{-\totalheight}{\includegraphics[width=0.20\textwidth]{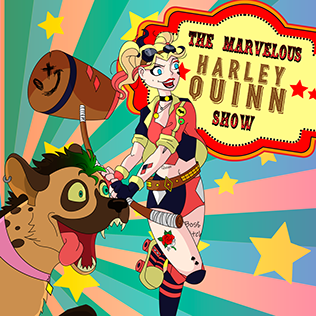}} & \vspace{0.55cm} \raisebox{-\totalheight}{\includegraphics[width=0.22\textwidth]{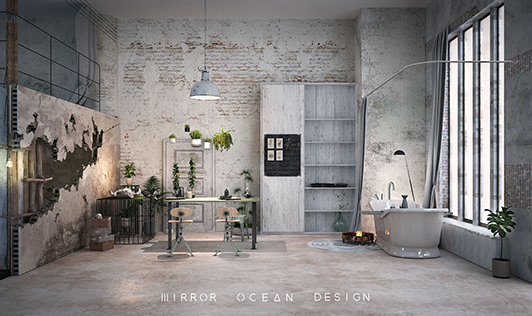}} & \vspace{-0.2cm} \raisebox{-\totalheight}{\includegraphics[width=0.18\textwidth,height=0.15\textheight]{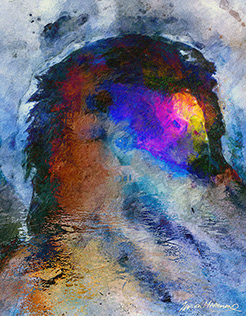}} 
            \\
            MS-COCO &
            a woman with a painting of a face on it &
            a close up of a remote with a remote &
            a kite that is hanging in the air & 
            a room with a lot of windows and a clock &
            a person riding a surfboard in the water \\
            
            \hline
            StyleBabel CL+IL &
            digital illustration of a fictional character using dark color tones against a black background & 
            product photography of business cards with text and logo in soft lighting against a white background & 
            digital illustration featuring animated characters and typography using bright colors & 
            architectural photography of a modern interior design in bright lighting using a neutral color palette & 
            abstract watercolor painting using paint brush strokes and shading effects
 
            \\
            
          \end{tabular}
      \end{adjustbox}
    \caption{Examples of natural language captions generated for various art styles; Generated captions are compared from VirTex models trained on MS-COCO and Stylebabel CL+IL, showing the benefit of the style information present in the dataset}
    \label{fig:senteneces_viz}
    \squeezeup
\end{figure*}


We run a user study on AMT to verify the correctness of the tags generated, presenting 1000 randomly selected test split images alongside the top tags generated for each. For each image/tags pair, 3 workers are asked to indicate tags that don't fit the image. We score tags as correct if all 3 workers agree they belong. The absolute accuracy of this study is 89.86\%, indicating high tag generation accuracy. 

\begin{figure}
    \begin{center}
        \includegraphics[width=\linewidth,height=4cm]{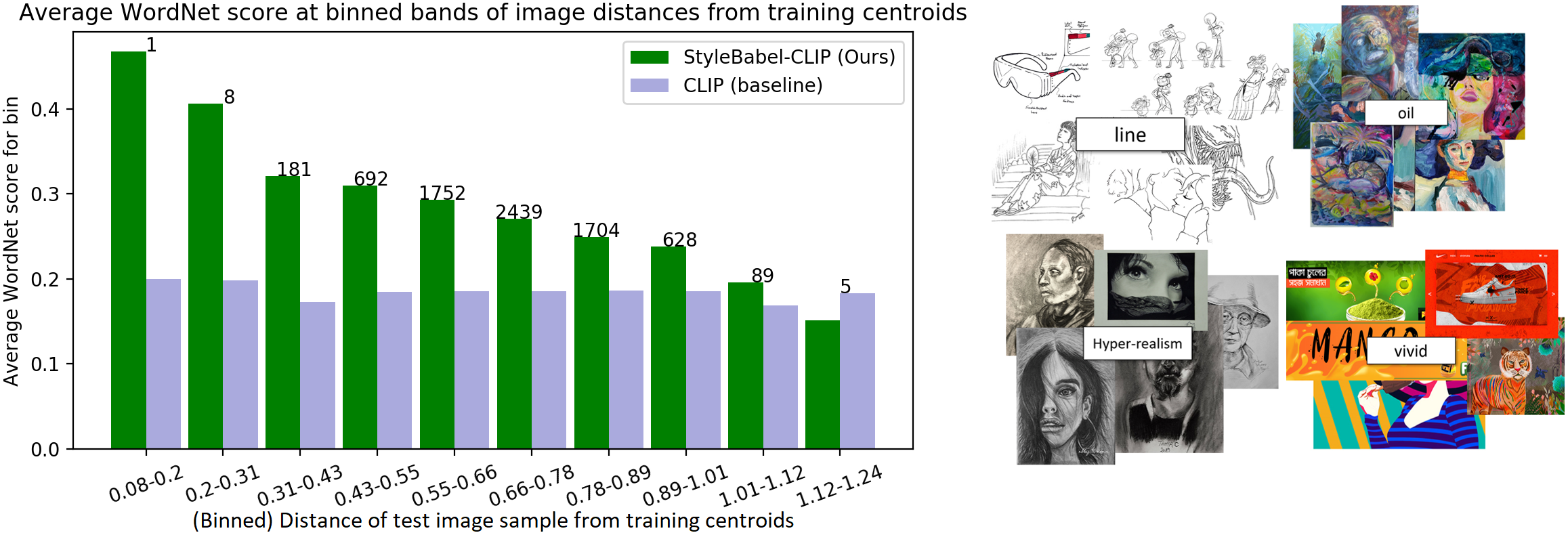}  \\
    \end{center}
    \squeezeup
    \caption{(left) Generalization experiment for tag generation using baseline CLIP \cite{clip} and our StyleBabel trained model using ALADIN-ViT. The test set was sorted by distance in the style embedding space to closest training cluster. Their WordNet scores were binned into 10 quantized distance bands. The numbers atop the bars indicate the number of samples in the corresponding bin. (right) Top 5 style retrieval using textual tags.  Using tags generated by ALADIN-ViT/CLIP over the StyleBabel test partition, we perform a keyword based search for artistic style.}
    \squeezeup
    \squeezeup
    \label{fig:generalisability}
\end{figure}

Finally, we explore the model's generalization to new styles by evaluating the average WordNet score of images from the test split. In Fig.\ref{fig:generalisability}, we group the data samples into 10 \textit{bins} of distances from their respective style cluster centroid, in the style embedding space. As before, we compute the WordNet score of tags generated using our model and compare it to the baseline CLIP model. Though the quality of the CLIP model is constant as samples get further from the training data, the quality of our model is significantly higher for the majority of the data split. At worst, our model performs similar to CLIP and slightly worse for the 5 most extreme samples in the test split. But for the majority of the test data, our model considerably outperforms CLIP. 

\begin{table*}[h!]
  \centering
  \small
  \begin{adjustbox}{width=0.9\textwidth}
      \centering
      \begin{tabular}{l|l|r|r|r|r|r|r|r}
        \hline
        Data                 & Model & Bleu-1 & Bleu-2 & Bleu-3 & Bleu-4 & METEOR & Rouge-L & CIDEr \\ 
        \hline
        MS-COCO baseline     & VirTex & 0.162 & 0.053 & 0.016 & 0.005 & 0.037 & 0.145 & 0.022 \\ 
        StyleBabel (CL)      & VirTex & 0.127 & 0.049 & 0.022 & 0.010 & 0.054 & 0.135 & 0.076 \\ 
        StyleBabel (IL)      & VirTex & 0.331 & 0.187 & 0.113 & 0.071 & 0.129 & 0.288 & 0.350 \\ 
        StyleBabel (CL+IL) & VirTex & \textbf{0.335} & \textbf{0.189} & \textbf{0.118} & \textbf{0.078} & \textbf{0.131} & \textbf{0.288} & \textbf{0.372} \\ 
        \hline
        StyleBabel (CL+IL) & ResNet LSTM       & 0.087 & 0.021 & 0.008 & 0.002 & 0.033  & 0.080 & 0.017 \\
        StyleBabel (CL+IL) & ALADIN-ViT LSTM   & 0.094 & 0.030 & 0.013 & 0.006 & 0.042  & 0.089 & 0.034 \\
        \hline
        VirTex & Artemis    & 0.185 & 0.083 & 0.041 & 0.023 & 0.081 & 0.182 & 0.146 \\
        VirTex & Artemis (SB)& 0.120& 0.031 & 0.013 & 0.005 & 0.034 & 0.108 & 0.029 \\
        \hline
      \end{tabular}
  \end{adjustbox}
    \caption{(top) VirTex caption generation metrics on 1k holdout StyleBabel test data. CL represents cluster-level, and IL represents image-level. We also run CL+IL, where we fine-tune a cluster-level model with image-level data labels. (middle) Results with a baseline LSTM model trained over either ResNet or ALADIN image embeddings (bottom) Additional experiments showing performance of a VirTex model trained on an existing dataset (Artemis), and also evaluated on the StyleBabel (SB) test set.}
    \label{tab:caption_metrics}
  \squeezeup
  \squeezeup
\end{table*}

\subsection{Style Description (style2text)}

\noindent We explore the feasibility of using the StyleBabel dataset for generating natural language  captions. We conduct our experiments using a fusion of the VirTex \cite{virtex} backbone for the visual representation learning of image/caption pairs, and Attention on Attention (AoA) \cite{aoa} for the caption decoding. VirTex encodes images without using scene graphs, therefore avoiding issues related to style not being localized in an image. VirTex replaces the Faster-RCNN \cite{faster-rcnn} component in AoA to generate feature maps. We use the pre-trained VirTex on COCO \cite{mscoco}, and fine-tune the entire setup, end-to-end on final StyleBabel captions. 
As per standard practice, during data pre-processing, we remove words with only a single occurrence in the dataset. Removing 45.07\% of unique words from the total vocabulary, or 0.22\% of all the words in the dataset.
We test the caption generation on the StyleBabel test data across Bleu \cite{bleu_score}, METEOR \cite{meteor_score}, Rouge \cite{rouge_score}, and CIDEr \cite{cider_score}, as shown in Tab \ref{tab:caption_metrics}. See the supplementary material for further analysis.

We would like to stress that these metrics are not comparable with values from these metrics on standard literature, as we are solving a new task. In literature, these metrics are used for semantic, localized features in images, whereas our task is to generate captions for global, style features of an image. We include an MS-COCO baseline, to show comparative accuracy versus a dataset with no style information. Figure \ref{fig:senteneces_viz} displays captions generated using this method.

\subsection{Text based style retrieval (text2style)}

\noindent We explore the potential of the ALADIN-ViT+CLIP model trained in subsec.~\ref{sec:clip_tags} to perform image retrieval, using textual tag queries. By first indexing the score assigned to each tag in the dictionary at the image level, we can then use a tag query to retrieve images based on the sorted scores for that tag. We use nearest-neighbour search using the image embeddings, reversing the tags generation experiment. Fig \ref{fig:generalisability} shows some example image retrievals using text queries. 

To measure the quality of the results, we run all text tags as queries. For each, we compute the WordNet similarity of the query text tag to the \textit{k}th top tag associated with the image, following a tag retrieval using a given image. We vary \textit{k} and collect the average scores at values of 1, 5, 10, and 25. The scores at these values are 0.72, 0.467, 0.392, and 0.332, respectively .


\section{Conclusion}

We proposed StyleBabel, a novel unique dataset of digital artworks and associated text describing their fine-grained artistic style. Our annotation approach was inspired by Grounded Theory (GT) \cite{ground},
to support the ‘emergence’ of themes from the corpus of digital artwork -- as opposed to fitting images to pre-existing categories \cite{charmaz}. These sessions generated discussion while simultaneously evolving and arriving through consensus at a shared vocabulary for describing image clusters of similar style.

We extended ALADIN~\cite{aladin} to incorporate a visual transformer \cite{vit_paper} (ALADIN-ViT) encoder, obtaining state of the art style similarity discrimination, leveraging StyleBabel for the automated description of artwork images using keyword tags and captions.  We also showed text-based image retrieval of images based on the generated style tags. Further work could explore use of tags as priors in generating captions, and exploring more downstream tasks using StyleBabel.

\section{Acknowledgement}
We would like to thank Thomas Gittings, Tu Bui, Alex Black, and Dipu Manandhar for their time, patience, and hard work, assisting with invigilating and managing the group annotation stages during data collection and annotation.

{\small
\bibliographystyle{ieee_fullname}
\bibliography{main}
}

\end{document}